\documentclass[journal]{IEEEtran}

\usepackage{lineno,hyperref}
\usepackage{CJK}
\usepackage[top=2cm, bottom=2cm, left=2cm, right=2cm]{geometry}
\usepackage{algorithm}
\usepackage{algorithmicx}
\usepackage{algpseudocode}
\usepackage{amsmath}
\usepackage{epstopdf}
\usepackage{subfigure}
\makeatletter
\def\set@curr@file#1{%
  \begingroup
    \escapechar\m@ne
    \xdef\@curr@file{\expandafter\string\csname #1\endcsname}%
  \endgroup
}
\def\quote@name#1{"\quote@@name#1\@gobble""}
\def\quote@@name#1"{#1\quote@@name}
\def\unquote@name#1{\quote@@name#1\@gobble"}
\makeatother
\usepackage{graphicx}

\modulolinenumbers[5]

\begin{document}

\title{Black Box Algorithm Selection by Convolutional Neural Network}

\author{Yaodong~He,~and~Shiu~Yin~Yuen,~\IEEEmembership{Senior~Member,~IEEE}
\thanks{Y. He and S.Y. Yuen are with the Department
of Electrical Engineering, City university of Hong Kong (e-mail: yaodonghe2-c@my.cityu.edu.hk; kelviny.ee@cityu.edu.hk).}
\thanks{\textit{(Corresponding author: Shiu Yin Yuen.)}}}

\maketitle

\begin{abstract}
Although a large number of optimization algorithms have been proposed for black box optimization problems, the no free lunch theorems inform us that no algorithm can beat others on all types of problems. Different types of optimization problems need different optimization algorithms. To deal with this issue, researchers propose algorithm selection to suggest the best optimization algorithm from the algorithm set for a given unknown optimization problem. Usually, algorithm selection is treated as a classification or regression task. Deep learning, which has been shown to perform well on various classification and regression tasks, is applied to the algorithm selection problem in this paper. Our deep learning architecture is based on convolutional neural network and follows the main architecture of visual geometry group. This architecture has been applied to many different types of 2-D data. Moreover, we also propose a novel method to extract landscape information from the optimization problems and save the information as 2-D images. In the experimental section, we conduct three experiments to investigate the classification and optimization capability of our approach on the BBOB functions. The results indicate that our new approach can effectively solve the algorithm selection problem.
\end{abstract}


\IEEEpeerreviewmaketitle

\section{Introduction}
\label{sec:intro}
\IEEEPARstart{M}{any} engineering problems can be treated as black box optimization problems \cite{roy2008recent,zhang2013novel}. The algebraic expression and gradient information of this type of problems are unknown. Thus, traditional numerical methods are not suitable for them. According to the number of objectives, black box optimization problems can be classified as single-objective and multi-objective problems. According to the type of variables, they can be classified as continuous and discrete problems. In this paper, without loss of generality, we only focus on continuous single-objective optimization problems.

Although a large number of optimization algorithms such as evolutionary algorithms have been invented, the no free lunch (NFL) theorems inform us that no algorithm can beat others on all optimization problems \cite{wolpert1997no}. Thus, we still need different algorithms for different types of problems. To deal with this issue, the algorithm selection problem is proposed. Given an unknown optimization problem, the goal of algorithm selection is to predict the most suitable algorithm from a pre-defined algorithm set. Usually, the predictor is a well-trained classification model or a regression model. Using a classification model or a regression model depends on whether one treats the algorithm selection problem as a classification task or a regression task. In the paper, we consider it as a classification task.

Although many researches have been done on algorithm selection for constraint satisfaction problems, black box algorithm selection has attracted little attention until the past decade \cite{kerschke2019aut}. Most of the researches focus on defining suitable features or using reliable machine learning models. In recent years, deep neural network, also called deep learning \cite{lecun2015deep}, and its variants have won many competitions in various fields including computer vision, natural language processing, speech recognition, etc. However, only a few works focus on applying deep learning models to the algorithm selection problem. To fill the gap, we propose a novel approach to extract landscape information from optimization problems, and use a convolutional neural network (CNN), which is a variant of deep neural network, to understand the information. In this paper, we do not use any existing human-defined landscape features.

This paper has three main contributions: 1) We propose a novel method to extract landscape information from optimization problems and save the information as 2-D images. One can use this method to extract landscape information from optimization problems for their own deep learning architecture. 2) We apply a well-known structure, which is called Visual Geometry Group (VGG) \cite{simonyan2014very} in the computer vision field, to algorithm selection. 3) We build an overall framework applying deep learning to algorithm selection problem and show its effectiveness.

The rest of this paper is organized as follows: We review and summarize related works in section \ref{sec:related_work}. The framework to extract landscape information and details of the applied convolutional neural network are introduced in section \ref{sec:methodology}. In section \ref{sec:exper}, we conduct three experiments to investigate the performance of our approach. In the last section, we conclude this paper and discuss the future works.

\section{Related Works}
\label{sec:related_work}
\subsection{Portfolio approaches and ensemble approaches}
Besides algorithm selection approaches, combining and running multiple optimization algorithms together is another way to solve the issue mentioned in NFL theorems. This type of approaches are called portfolio approaches or ensemble approaches. So far, many portfolio approaches have been proposed. Peng et al. \cite{peng2010population} propose a population based portfolio approach. In their approach, two parameters \textit{l} and \textit{s} are pre-defined. For each \textit{l} generations, the worst \textit{s} candidates generated by each algorithm are selected and replaced by the best \textit{s} candidates generated by other algorithms. The budget is equally shared by the algorithms from the algorithm set. Vrugt et al. \cite{vrugt2008self} propose an approach called multi-algorithm genetically adaptive method. At the end of each generation, the number of offsprings generated by each algorithm is modified according to its previous performance. Yuen et al. \cite{yuen2016algorithm} propose a portfolio approach called MultiEA. At the end of each iteration, this approach will predict the fitness of each algorithm at the nearest common future point. The algorithm that has the best predicted fitness will be run for the next iteration. 

Different from the approaches mentioned in the last paragraph, there is another type of approaches that combine the same algorithm with different parameter settings. Zhao et al. \cite{zhao2012decomposition} propose a decomposition-based multiobjective evolutionary algorithm which uses ensemble of neighbourhood sizes. Wu et al. \cite{wu2016differential} combine multiple mutation strategies to improve the performance of differential evolution variants. Three different mutation strategies are applied. A detail survey paper about portfolio and ensemble approaches can be found in \cite{wu2019ensemble}.

\subsection{Exploratory landscape analysis features and Algorithm selection approaches}
In the past two decades, most researchers focus on solving the algorithm selection problem by conventional machine learning models. These models are not as complicated as deep neural network but require well-defined features, or else their performance will not be gratifying. These human-defined features for algorithm selection are called exploratory landscape analysis (ELA) features \cite{munoz2015algorithm}. Now, a large number of ELA features have been proposed. Some examples are fitness distance correlation, probability of convexity, dispersion, entropy, etc. These features can describe landscape characteristics such as separability, smoothness, basins of attractions, ruggedness, etc. Kerschke et al. \cite{kerschke2017flacco} release a user platform to calculate these ELA features. Researchers can apply these features to different machine learning models. Based on their own platform, Kerschke et al. \cite{kerschke2019automated} use various machine learning models including support vector machine, regression tree, random forest, extreme gradient boosting and multivariate adaptive regression splines for classification task, regression task, and pairwise regression task. Bischl et al. \cite{bischl2012algorithm} treat algorithm selection as a regression task and propose a cost sensitive learning method to predict the performance of optimization algorithms on the black-box optimization benchmarking (BBOB) functions \cite{hansen2010comparing}. Mu{\~n}oz et al. \cite{munoz2014exploratory} propose an information content-based method for algorithm selection problem. The classification accuracy of their method which classifies the BBOB function into five groups is higher than 90$\%$. In \cite{munoz2017performance}, Mu{\~n}oz et al. use footprints to analyze the performance of algorithms on the BBOB functions. They cluster combinations of algorithms and problems into different regions. These regions can depict which algorithm is statistically better than others on which problem. The result can be used to motivate researchers to design new algorithms for unexplored regions. He et al. \cite{he2019sequential} propose a sequential approach. The approach incorporates a restart mechanism. At the beginning of each restart, a set of algorithm based ELA features \cite{he2018exploratory} are extracted and used as the input to prediction models. The models will suggest an optimization algorithm for the next run. The optimization will not stop until the maximal number of evaluations are reached. A recently published survey paper about algorithm selection approaches can be found in \cite{kerschke2019aut}.

\subsection{Deep learning and its application to algorithm selection problem}
Deep neural network, also called deep learning, has shown great capability of solving classification and regression tasks in various engineering fields. Compared with conventional machine learning models, it requires less human-defined features but many training instances. This method is good at learning from hidden information. However, only a few researches focus on applying deep learning to algorithm selection. In \cite{munoz2012meta}, Mu{\~n}oz et al. try to predict the performance of CMA-ES by using deep neural network. Their network has two hidden layers and 10 trainable parameters for each hidden layer. A set of ELA features and algorithm parameters are concatenated as the input. However, the input size and the number of trainable parameters are small. The approach cannot completely show the capability of deep neural network. Loreggia et al. \cite{loreggia2016deep} use convolutional neural network to recommend optimization algorithms for the boolean satisfiability (SAT) problem and show appealing results. Since the SAT problems are described in text files, they convert these text files to images. The images are the input of a convolutional neural network with three convolutional layers and two fully connected layers. However, the authors did not propose an approach to solve the optimization problems that cannot be described in text files. 

\begin{figure*}[h!]
\centering
\includegraphics[width=\textwidth, height=10cm]{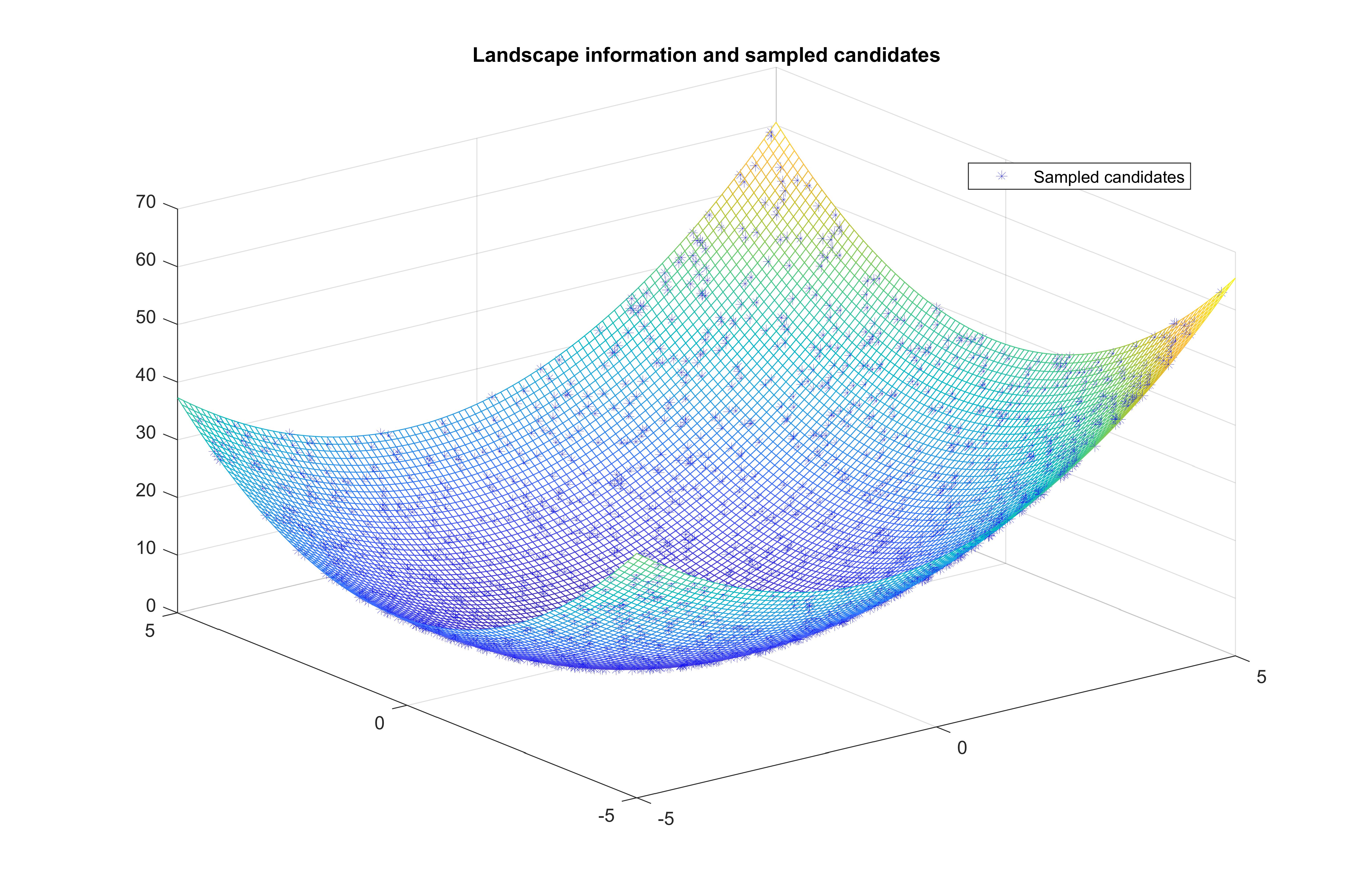}
\caption{Landscape information and sampled candidates.}
\label{sample}
\end{figure*}

\begin{figure*}[h!]
\centering
\includegraphics[width=\textwidth, height=15cm]{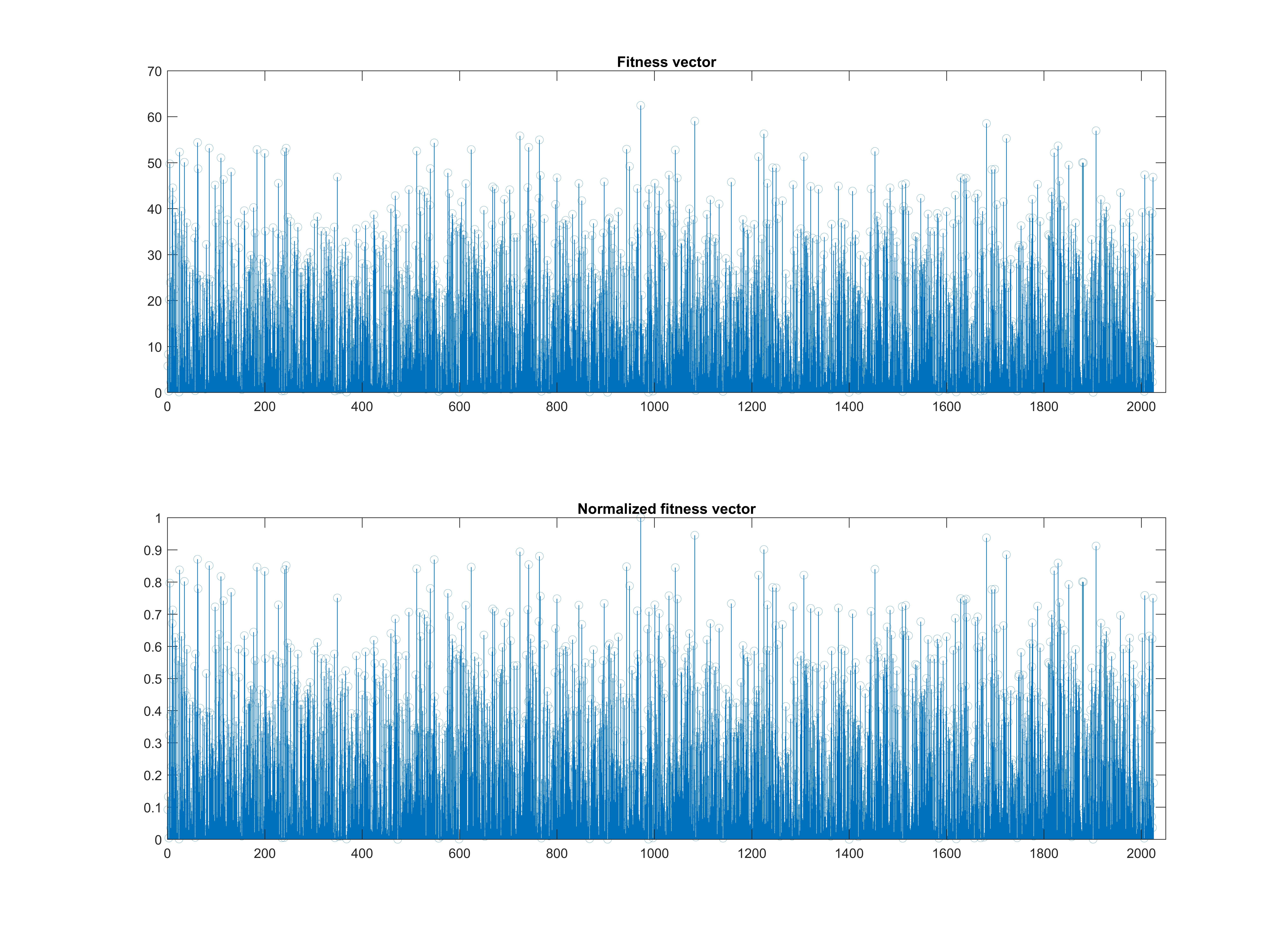}
\caption{Fitness vector and normalized fitness vector.}
\label{vector}
\end{figure*}

\begin{figure*}[h!]
\centering
\includegraphics[width=\textwidth, height=13cm]{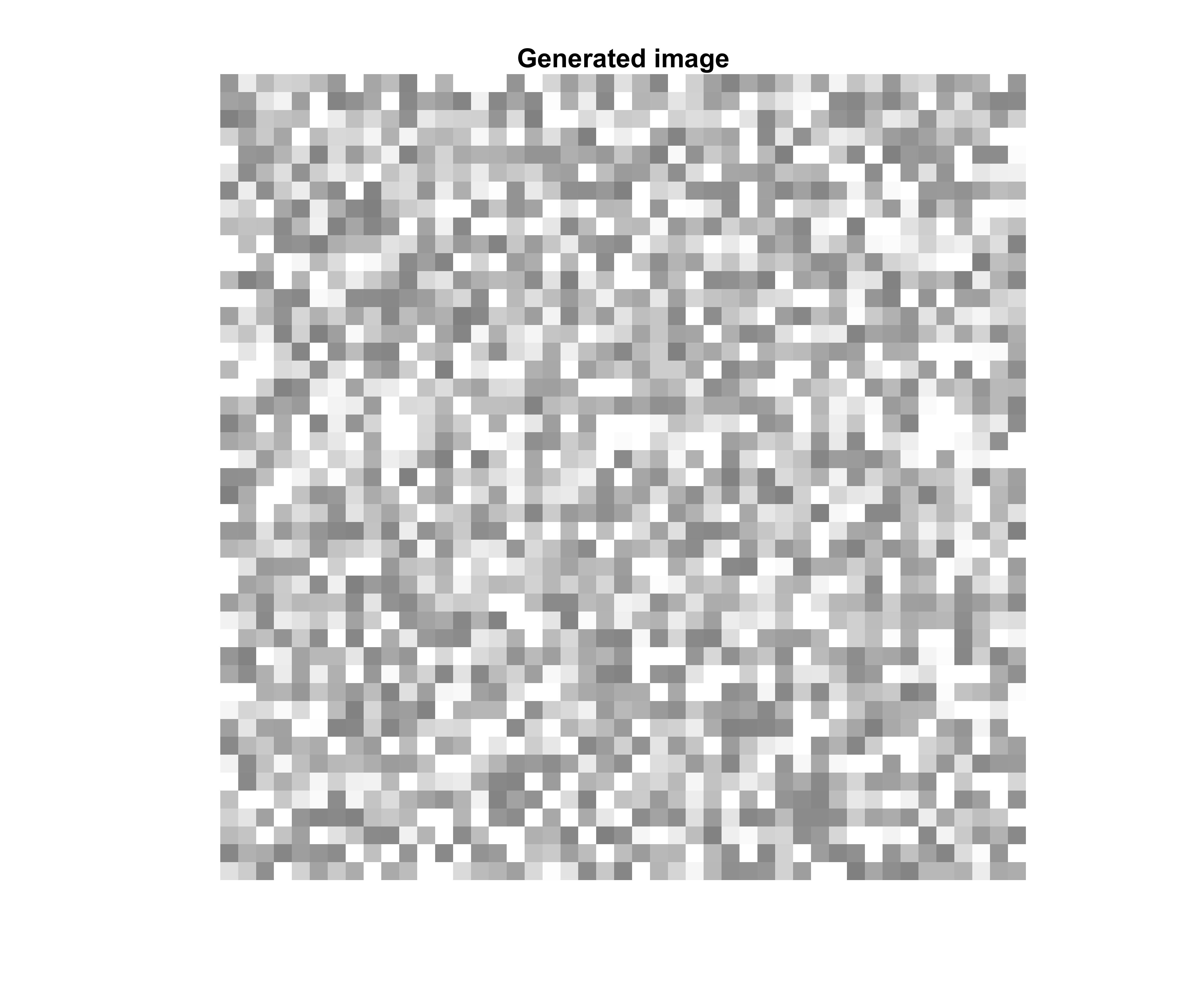}
\caption{Generated 2-D image.}
\label{image}
\end{figure*}

\section{Algorithm selection by convolutional neural network}
\label{sec:methodology}
In this section, we introduce our overall framework. In the first subsection, we introduce a novel method to extract landscape information from the optimization problems and save the information as images. In the second subsection, we introduce the architecture of our network.
\subsection{Extract information from optimization problems}
The process of sampling landscape information and saving the information as images from optimization problems is described as follows:

1) We use scaling methods to make the search space of all training and testing instances the same. In the experimental section, we use the BBOB function instances which have an identical search space $[-5, 5]^{D}$ as our training and testing instances. Thus, they do not need scaling. 

2) We uniformly generate a set of sample coordinates from the entire search space. The generated sample coordinates are recorded as a $N\times D$ matrix, where $N$ is the number of sampled candidates and $D$ is the problem dimension.

3)	Given an optimization problem, according to the $N\times D$ matrix generated in 1), we sample a set of \textit{N} candidates from the entire search space for the problem. The fitness results of the candidates compose a \textit{N}-length fitness vector $\overrightarrow{Fit}$. The $i^{th}$ value of $\overrightarrow{Fit}$ records the $i^{th}$ fitness result of the sampled candidates.

4)	The sampled fitness results are normalized to [0, 1] by the following equation.

\begin{equation}
\overrightarrow{Fit}_{normalized}=\dfrac{\overrightarrow{Fit}-MIN(\overrightarrow{Fit})}{MAX(\overrightarrow{Fit})-MIN(\overrightarrow{Fit})}
\end{equation}

where $MAX$ and $MIN$ operations are to choose the maximal and minimal value of the given vector $\overrightarrow{Fit}$ respectively.

5)	We resize the normalized fitness vector $\overrightarrow{Fit}_{normalized}$ to a $\sqrt{N}\times\sqrt{N}$ 2-D image by standard image resize operators.

We give an example to illustrate how an image is generated. Suppose the given optimization problem is a 2-D problem. Note that our approach can solve higher dimensional problems. Using 2-D problem in this example is for a better visualization. Fig. \ref{sample} shows the landscape of the optimization problem and the sampled candidates. We uniformly sample 2025 candidates from the entire search space. The fitness results of the 2025 candidates compose a fitness vector. Fig. \ref{vector} shows the fitness vector $\overrightarrow{Fit}$ and the normalized fitness vector $\overrightarrow{Fit}_{normalized}$. The normalized fitness vector is converted to a $45\times 45$ image. The image is shown as a gray scale image in Fig. \ref{image}.

\subsection{Convolutional neural network}
Our CNN architecture follows the VGG architecture. Compared with other CNN architectures, VGG incorporates a smaller kernel size and a greater depth. In the experimental section, we investigate the problem with dimension $D=2$ and $D=10$. The sample size of the two types of problems should be different. We set the sample size of problem with $D=2$ and $D=10$ to 2025 and 10000 respectively. Since the input sizes of the two types of problems are different, two slightly different CNN architectures (a) and (b) are presented in this section.

The overall architecture consists of several groups of convolutional layers, several fully connected layers and a softmax layer. We use a softmax layer as the output layer because we treat the algorithm selection problem as a classification task, and the softmax layer is a commonly used output layer for the classification task. The number of outputs equals to the number of classes. We use rectified linear unit (ReLU) as the activation function and the max pooling filter to reduce the size of feature map.

We adopt ReLU at the end of each layer. ReLU is a commonly used activation function which retains the non-negative part and forces the negative part to be 0. It has been shown performing well on 2-D data \cite{glorot2011deep}. The formula of ReLU can be described as $f(x) = MAX(0, x)$, where $x$ is the input of ReLU and $f(x)$ is the output of ReLU. 

We also adopt max pooling layers at the end of each group layer. These pooling layers are used to reduce the size of the output feature maps, which are the input of the next layers.

\begin{figure*}[h!]
\centering
\includegraphics[width=\textwidth, height=11cm]{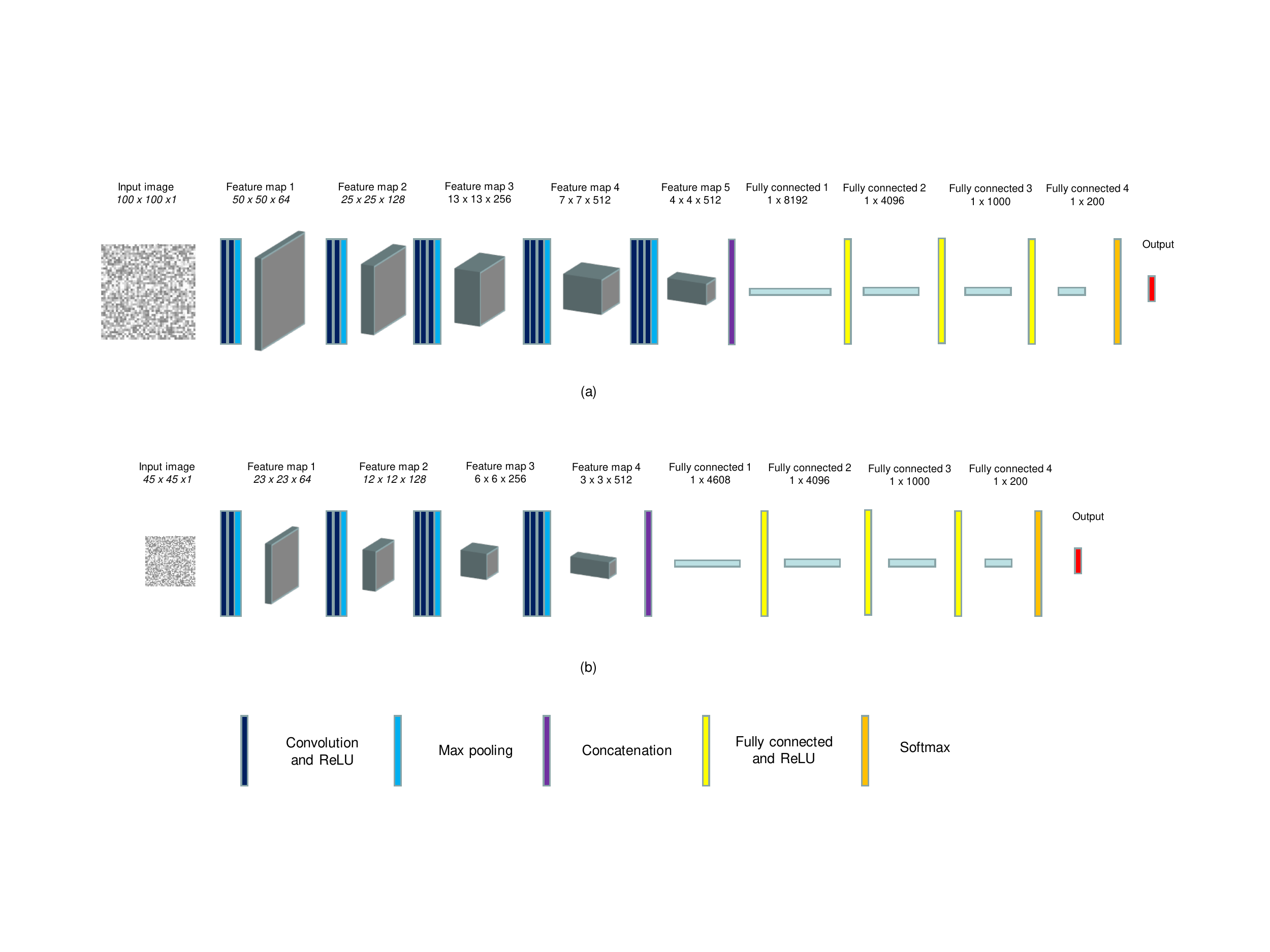}
\caption{The overall CNN architecture.}
\label{framework}
\end{figure*}

The overall framework is shown in Fig \ref{framework}. The detailed settings of the architecture is shown in Table \ref{par1}. In this paper, we provide two convolutional architectures for two different types of input data. The architecture (a) for the instances with $D=10$ consists of five groups of convolutional layers, three fully connected layers and a softmax layer. The convolutional layers are identical to those in VGG16. The architecture (b) for the instances with $D=2$ consists of four groups of convolutional layers, three fully connected layers and a softmax layer. The four groups of convolutinal layers of the architecture (b) are identical to the first four groups of convolutional layers in the architecture (a).

\begin{table}[!h]

\caption{Parameter settings of the architectures. The convolutional group 5 is not included in the architecture for the instances with $D=2$.}

\centering  
\begin{tabular}{ p{1.2cm}p{1.6cm} p{1.2cm} p{1cm} p{1.5cm}} 
\hline
group&layers & kernel size &stride&output channel \\
\hline
group 1&conv3-64 & 3$\times$3  &1&64 \\
&conv3-64 & 3$\times$3  &1&64 \\
&max2 & 2$\times$2  &2&64 \\
\hline
group 2&conv3-128 & 3$\times$3  &1&128 \\
&conv3-128 & 3$\times$3  &1&128 \\
&max2 & 2$\times$2  &2&128 \\
\hline
group 3&conv3-256 & 3$\times$3  &1&256 \\
&conv3-256 & 3$\times$3  &1&256 \\
&conv3-256 & 3$\times$3  &1&256 \\
&max2 & 2$\times$2  &2&128 \\
\hline
group 4&conv3-512 & 3$\times$3  &1&512 \\
&conv3-512 & 3$\times$3  &1&512 \\
&conv3-512 & 3$\times$3  &1&512 \\
&max2 & 2$\times$2  &2&512 \\
\hline
group 5&conv3-512 & 3$\times$3  &1&512 \\
&conv3-512 & 3$\times$3  &1&512 \\
&conv3-512 & 3$\times$3  &1&512 \\
&max2 & 2$\times$2  &2&512 \\
\hline
&fc-4096 & 1$\times$1  &1&4096 \\
&fc-1000 & 1$\times$1  &1&1000 \\
&fc-200 & 1$\times$1  &1&200 \\
&softmax & 1$\times$1  &1&number of classes \\
\hline
\end{tabular}
\label{par1}
\end{table}

The number of outputs of softmax layer equals to the number of classes. For example, if the task is to recommend a suitable algorithm for the given problem from the algorithm set consisting of three algorithms, the number of outputs of softmax layer in this task should be three. Softmax layer use cross entropy loss to evaluate the classification error. The formula of cross entropy loss is:
\begin{equation}
Loss=\dfrac{1}{N}\sum_{i=1}^{N}-p_{i}{\rm log}(\hat{p_{i}})-(1-p_{i}){\rm log}(1-\hat{p_{i}})
\label{lossc}
\end{equation}
where $p_{i}$ is the actual probability distribution of the $i^{th}$ instance, and $\hat{p_{i}}$ is the predicted probability distribution of the $i^{th}$ instance. $N$ is the number of instances used for each training step. Our CNN architecture is to minimize the loss value $Loss$.
\section{Experimental results}
\label{sec:exper}
\subsection{Data description}
In the following experiments, we use the BBOB functions to evaluate the performance of our approach. The BBOB functions can be classified into 24 problem classes. For each class, different problem instances can be generated by random transformation and rotation. The search space of the instances are $[-5, 5]^{D}$, where $D$ is the problem dimension. We investigate instances with $D=2$ and $D=10$ in the first experiment, and instances with $D=10$ in the second and the last experiment.

\subsection{Training details}
We set the number of training epochs to 150, where one epoch denotes the network is trained by the entire training data once. We set the batch size, which is the number of instances for each training, to 60. We use the Adam optimizer \cite{kingma2014adam} to optimize the neural network, and set the learning rate to $10^{-4}$. The first momentum, the second momentum and epsilon of the Adam optimizer is set to 0.9, 0.999 and $10^{-8}$ respectively following their recommended settings. The CNN architecture is implemented by Python language with Tensorflow library on a Nvidia 1080Ti graphic card. At the end of each epoch, we use the validation instances to test the performance of our network. The network whose parameters perform the best (i.e., having the highest accuracy) on the validation instances will be saved. After the training, we use the testing instances to evaluate the performance of the well-trained network. To obtain reliable results, we repeat the training and testing process five independent times and record the mean of the prediction accuracy. 

\subsection{Classification results of classifying problem instances to different classes}
In the first experiment, we investigate the classification capability of our approach on the BBOB function instances with $D=2$ and $D=10$. In the BBOB competition settings \cite{hansen2010comparing}, the total number of evaluations for an optimization process is $10000\times D$, where $D$ is the problem dimension. In our experiment, the number of sampled candidates for the two kinds of instances are approximately $10\%$ of the total budget. We sample $45\times45=2025$ candidates for instances with $D=2$ which corresponds to $10.125\%$ of the total budge, and sample $100\times100=10000$ candidates for instances with $D=10$ which corresponds to $10\%$ of the total budge. We investigate the instances with $D=2$ because their landscape are similar to images. We investigate instances with $D=10$ because this kind of instances are commonly investigated in many papers of algorithm selection field.

In this experiment, we evaluate the performance of using the deep neural network to classify the BBOB functions. For each of the 24 problem class, we generate 250 different instances. Thus, there are $250\times24=6000$ instances in total. We use $70\%$ instances for training, $10\%$ instances for validation, and $20\%$ instances for testing. Given an unknown instance from the testing set, the classification task for our approach is to predict which problem class it belongs to. There are 24 possible prediction results in total. Thus, the baseline accuracy should be $1/24=4.17\%$. Table \ref{acc1} shows the classification results for instances with $D=2$ and $D=10$.

\begin{table}[!h]

\caption{The accuracy of classifying the BBOB instances with $D=2$ and $D=10$ into 24 problem classes.}
\centering  
\begin{tabular}{ p{3cm}p{2cm} p{2cm} } 
\hline
&$D=2$ & $D=10$  \\
\hline
f1	&100.00$\%$	&100.00$\%$\\
f2	&65.20$\%$	&	59.20$\%$\\	
f3	&90.80$\%$	&	82.40$\%$\\	
f4	&51.20$\%$		&54.80$\%$\\	
f5	&100.00$\%$	&	87.20$\%$\\	
f6	&66.80$\%$	&	26.80$\%$\\	
f7	&31.60$\%$	&	37.60$\%$\\	
f8	&93.20$\%$	&	90.40$\%$\\	
f9	&98.40$\%$	&	96.40$\%$\\	
f10	&18.40$\%$	&	22.40$\%$\\	
f11	&25.60$\%$	&	31.60$\%$\\	
f12	&44.40$\%$		&14.00$\%$\\	
f13	&89.20$\%$	&	59.20$\%$\\	
f14	&33.60$\%$		&47.60$\%$\\	
f15	&52.00$\%$	&	50.00$\%$\\	
f16	&56.80$\%$		&69.20$\%$\\	
f17	&49.60$\%$		&21.20$\%$\\	
f18	&45.60$\%$		&37.20$\%$\\	
f19	&99.60$\%$	&	76.00$\%$\\	
f20	&98.00$\%$	&	94.80$\%$\\	
f21	&32.00$\%$	&	59.20$\%$\\	
f22	&38.40$\%$	&	22.40$\%$\\	
f23	&97.60$\%$	&	96.80$\%$\\	
f24	&88.40$\%$		&45.60$\%$\\	
\hline
Average	&65.27$\%$&	57.58$\%$\\	
\hline
\end{tabular}
\label{acc1}
\end{table}

Both results for the instances with $D=2$ and $D=10$ are much higher than the baseline accuracy. It indicates that our approach can classify the optimization functions into their corresponding classes. For the instances with $D=2$, the average accuracy is $65.27\%$. For the instances with $D=10$, the average accuracy is $57.58\%$. It is reasonable that the performance on the 2-D problems is better. According to \cite{morgan2013sampling}, if the increasing of the sample size does not match the exponential growth of $D$, the sample strategy will suffer from a convergence behaviour. This convergence behaviour discussed in depth in \cite{morgan2013sampling}. However, in our experiment, the sample size increases linearly. The phenomenon also implies that the deep neural network may not work for very high dimensional optimization problems unless a large sample size is applied.

For the problem instances having salient characteristics such as Sphere function (which is f1 of the BBOB suite), our approach can identify them into correct classes with a very high accuracy. However, for the problem instances having no salient characteristics such as Ellipsoidal function (both f2 and f10 of the BBOB suite belong to Ellipsoidal function), our approach inevitably classifies many instances into incorrect classes. Misclassification happens because of the inherent similarity between functions rather than the limitations of our approach. This may not be a big problem as problem instances sharing similar characteristics may map to the same optimization algorithm.

\subsection{Classification results of algorithm selection}
In this experiment, we investigate the performance of our approach on recommending the optimization algorithms for the BBOB functions with $D=10$. The algorithm set consists of three optimization algorithms. They are artificial bee colony (ABC) \cite{karaboga2007powerful}, ovariance matrix adaptation evolution strategy (CMA-ES) \cite{hansen2003reducing} and linear population size reduction differential evolution (L-SHADE) \cite{tanabe2014improving}. The population sizes of ABC and CMA-ES are 125, 40 respectively. The initial population size of L-SHADE is set to 200. The settings follow their recommended settings in \cite{karaboga2007powerful, hansen2003reducing, tanabe2014improving}.

\begin{table}[!h]

\caption{The accuracy of recommending the most suitable algorithms for the BBOB instances with $D=10$.}
\centering  
\begin{tabular}{ p{3cm}p{2cm} } 
\hline
 & $D=10$ \\
\hline
f1	&100.00$\%$\\
f2	&95.40$\%$\\
f3	&95.60$\%$\\
f4	&94.80$\%$\\
f5	&92.40$\%$\\
f6	&98.00$\%$\\
f7	&64.40$\%$\\
f8	&89.20$\%$\\
f9	&99.60$\%$\\
f10	&99.60$\%$\\
f11	&92.40$\%$\\
f12	&96.80$\%$\\
f13	&84.00$\%$\\
f14	&30.80$\%$\\
f15	&95.60$\%$\\
f16	&92.40$\%$\\
f17	&95.40$\%$\\
f18	&94.80$\%$\\
f19	&92.00$\%$\\
f20	&92.40$\%$\\
f21	&90.40$\%$\\
f22	&71.60$\%$\\
f23	&92.00$\%$\\
f24	&94.40$\%$\\
\hline
Average	&89.33$\%$\\
\hline
\end{tabular}
\label{acc2}
\end{table}

Different from the first experiment, we do not test on the instances with $D=2$ in this experiment. The reason is that the instances with $D=2$ are very easy to solve. For most instances with $D=2$, all the three algorithms can find the global optima. In this case, the algorithm selection problem is insignificant. We test on the instances with $D=10$ only. Following the previous settings, we also sample 10000 candidates and convert the sampled vector to $100\times100$ image for each problem instance with $D=10$.

For each of the 24 problem class, we also generate 250 different instances. Thus, there are 6000 instances in total. For each instance, we run the three optimization algorithms on it 51 independent times and select the best algorithm as the label. We still use $70\%$ instances for training, $10\%$ instances for validation, and $20\%$ instances for testing. However, there are still some instances that more than one algorithm can find the global optima. In this experiment, we eliminate these instances from our dataset because these instances are not suitable for training and testing. Using instances with undetermined label will aggravate the performance of our neural network. There are 991 instances marked with undetermined labels from the training data, 149 instances from the validation data, and 292 instances from the testing data. After eliminating these unsuitable instances, there are 3209 instances used for training, 451 instances for validation, and 908 instances for testing.

\begin{table}[!h]

\caption{The rank results of our deep learning approach and single optimization algorithms. In this table, (+), (-) and (=) denote the corresponding algorithm is inferior to, superior to, and equally well to the deep learning approach on the corresponding problem respectively.}
\centering  
\begin{tabular}{ p{1cm}p{1.3cm} p{1.3cm}p{1.3cm}p{1.3cm} } 
\hline
&Deep Learning& ABC &CMA-ES& L-SHADE\\
\hline
f1	&1	& 1(=) 	& 1(=) 	& 1(=) \\
f2	&1	& 1(=)& 	 1(=) 	& 1(=) \\
f3	&1	& 1(=) &	 4(+) &	 3(+) \\
f4&	1	& 1(=) &	 4(+) 	& 3(+) \\
f5	&1	& 1(=) &	 1(=) &	 1(=) \\
f6	&1	 &4(+) &	 1(=) 	& 1(=) \\
f7	&1	 &4(+) 	& 3(+) &	 1(=) \\
f8	&1	 &4(+) 	& 1(=) 	& 1(=) \\
f9	&1	 &4(+) 	& 1(=) 	& 1(=) \\
f10	&1	 &4(+) 	& 1(=) 	 &1(=) \\
f11	&1	 &4(+) 	& 1(=) 	 &1(=) \\
f12	&1	 &4(+) 	& 1(=) 	 &3(+) \\
f13	&1	 &4(+) 	& 1(=) 	 &3(+) \\
f14	&1	 &4(+) 	& 2(=) 	 &1(=) \\
f15	&2	 &4(+) 	& 1(=) 	 &3(+) \\
f16	&2	 &3(+) 	& 1(=) 	 &4(+) \\
f17	&1	 &4(+) &3(+) &2(=)\\ 
f18	&3	 &4(+) 	& 2(=) 	& 1(=) \\
f19	&2	 &4(+) &	 1(=) 	& 3(+)\\ 
f20	&3	 &1(-) 	& 4(+) 	& 2(-) \\
f21	&3	 &1(=) 	& 4(+) 	& 2(=) \\
f22	&3	 &1(=) 	 &4(+) 	& 2(=) \\
f23	&3	 &2(=) 	& 1(-) 	 &4(+) \\
f24	&2	 &4(+) 	& 1(=) 	 &3(+)\\
Average&1.583	&2.875	&1.875	&2.000\\
\hline
\hline
\end{tabular}
\label{rank1}
\begin{tabular}{ p{2.7cm}p{1.3cm} p{1.3cm}p{1.3cm} } 
& ABC &CMA-ES& L-SHADE\\
\hline
Inferior to the deep learning approach&15&7&9\\

Superior to the deep learning approach&1&1&1\\

Equally well to the deep learning approach&8&16&14\\
\hline
\end{tabular}

\end{table}

\begin{table}[!h]

\caption{The detailed average results of our deep learning approach and single optimization algorithms.}
\centering  
\begin{tabular}{ p{1cm}p{1.3cm} p{1.3cm}p{1.3cm}p{1.3cm} } 
\hline
&Deep Learning& ABC &CMA-ES& L-SHADE\\
\hline
f1&	0.00E+00&	0.00E+00	&0.00E+00&	0.00E+00\\
f2&	0.00E+00&	0.00E+00&	0.00E+00	&0.00E+00\\
f3&	0.00E+00&	0.00E+00	&4.88E+00&	5.15E-01\\
f4	&0.00E+00&	0.00E+00&	9.73E+00	&1.42E+00\\
f5&	0.00E+00&	0.00E+00	&0.00E+00&	0.00E+00\\
f6&	0.00E+00&	2.42E-01&	0.00E+00&	0.00E+00\\
f7	&0.00E+00&	1.37E+00&	5.15E-02&	0.00E+00\\
f8&	0.00E+00&	4.71E-03	&0.00E+00&	0.00E+00\\
f9	&0.00E+00&	9.58E-01&	0.00E+00&	0.00E+00\\
f10	&0.00E+00&	1.43E+03&	0.00E+00&	0.00E+00\\
f11	&0.00E+00&	3.12E+01&	0.00E+00&	0.00E+00\\
f12&	0.00E+00	&1.21E+00&	0.00E+00&	4.03E-05\\
f13&	0.00E+00&	6.61E-01&	0.00E+00&	1.76E-08\\
f14	&0.00E+00	&1.80E-03	&0.00E+00	&0.00E+00\\
f15	&3.65E+00	&2.19E+01	&2.96E+00&	6.45E+00\\
f16&	8.77E-01&	1.45E+00&	1.36E-01&	2.53E+00\\
f17	&1.79E-06&	1.21E+00	&8.00E-06&	6.96E-06\\
f18&	8.68E-04&	3.13E+00	&6.28E-04&	3.33E-05\\
f19	&4.29E-01&	1.38E+00	&1.19E-01&	1.36E+00\\
f20	&1.22E+00&	1.91E-01	&1.56E+00&	5.01E-01\\
f21	&1.68E-01&	1.42E-02	&2.54E+00&	1.65E-01\\
f22&	2.51E+00&	2.45E-01&	6.16E+00&	2.06E+00\\
f23	&9.18E-01&	4.76E-01	&2.43E-01&	1.10E+00\\
f24	&1.68E+01	&2.71E+01&	1.16E+01&	2.66E+01\\

\hline
\end{tabular}
\label{detailed1}
\end{table}

There are three possible predicted outputs. Thus, the baseline accuracy in this experiment should be $1/3=33.33\%$. Following the previous settings, we also record the parameters of the networks that perform the best on the validation instances and use testing instances to evaluate the recorded network. The training and testing process are also repeated five times to obtain a reliable result.

Table \ref{acc2} records the accuracy results. The average accuracy is $89.33\%$, which greatly outperforms the baseline accuracy. For most problem classes, the accuracy is more than 90$\%$. It indicates that our approach can recommend the most suitable optimization algorithms for most of the optimization problem instances.

\subsection{Comparisons of fitness results}
In this experiment, we investigate the optimization capability of our approach. We treat our approach as a portfolio approach and compare it with the three single optimization algorithms (i.e., ABC, CMA-ES and L-SHADE). We also test on the instances with $D=10$ only. For each instance, we use $10\%$ budget to sample candidates (i.e., $100\times100=10000$ candidates are sampled) and generate an image as the input of deep learning. In the second experiment, we repeatedly train and test five times and generated five well-trained networks. In this experiment, we use the well-trained network having the median accuracy performance in the second experiment to recommend an optimization algorithm from the algorithm set for each testing instance. We use the remaining $90\%$ budget to run the recommended algorithm on the instance. The fitness results found by our approach are compared with the results obtained by the single optimization algorithms with $100\%$ budget. Our approach and the three single optimization algorithms are run on each testing instance 51 independent times.

Table \ref{rank1} shows the rank results. We use Kruskal-Wallis test and multiple comparison test with p-value = 0.05 to test the significance. Compared with the three single algorithms, our deep learning approach has the best average ranks. For most problems, the deep learning approach can find the best optimization algorithm. Note that the key of our approach is to select the most suitable algorithm from the algorithm set. Thus, it is reasonable that our approach does not beat all algorithms on all problems. The detailed results with averaged fitness results are shown in table \ref{detailed1}.

\section{Conclusion and future works}
\label{sec:con}
Algorithm selection is an important topic since different types of optimization problems require different optimization algorithms. In recent years, researchers treat algorithm selection as a classification or regression task and propose many approaches for tackling it. On the other hand, deep neural network especially convolutional neural network has shown great capability of dealing with classification and regression tasks. However, only a few works focus on applying deep learning to algorithm selection, and little work has been done on applying deep learning to black box algorithm selection. In this paper, we propose an approach employing convolutional neural network architecture for algorithm selection.

We conduct three experiments to investigate the performance of our approach. In the first experiment, we show the classification capability of classifying the BBOB instances into their corresponding problem classes. In the second experiment, we show the capability of recommending optimization algorithms for given optimization problems. The accuracy is vastly better than the baseline accuracy. In the last experiment, we investigate the optimization capability of our approach. We use the well-trained convolutional neural network in the second experiment to recommend optimization algorithms, and use the predicted algorithms to optimize the given problems. The comparison results indicate that our deep learning approach can solve optimization problems effectively.

There are still lots of works that can be done in the future. In this paper, we only treat algorithm selection problem as a classification task, while treating it as a regression task is also reasonable. Moreover, our convolutional neural network follows the architecture of VGG, while other powerful architectures can also be used. The sampling strategy can also be improved. We uniformly sample candidates from the entire search space, while other sample strategies such as Latin hypercube sampling \cite{stein1987large} can also be applied.

\ifCLASSOPTIONcaptionsoff
  \newpage
\fi

\bibliographystyle{IEEEtran}  
\bibliography{mybib}

\end{document}